%% file: paper.tex
\documentclass[11pt]{article}
\usepackage[utf8]{inputenc}
\usepackage[T1]{fontenc}
\usepackage{amsmath,amssymb,amsthm}
\usepackage{graphicx}
\usepackage{hyperref}
\usepackage[margin=1in]{geometry}
\usepackage{booktabs}

\title{Statistical Parsing for Logical Information Retrieval}
\author{Greg Coppola, PhD}
\date{February 12, 2026}
\begin{document}
\maketitle
\input{sections/01_abstract}
\input{sections/02_introduction}
\input{sections/02b_motivation}
\input{sections/02c_bitter_lesson}
\input{sections/03_background}
\input{sections/04_language}
\input{sections/05_inference}
\input{sections/06_parsing}
\input{sections/07_llm_assisted}
\input{sections/08_experiments}
\input{sections/09_related_work}
\input{sections/11_conclusion}
\input{sections/11b_acknowledgments}
\input{sections/12_bibliography}
\end{document}

%% file: sections/01_abstract.tex
\begin{abstract}
In previous work (Coppola, 2024) we introduced the Quantified Boolean Bayesian Network (QBBN), a logical graphical model that implements the forward fragment of natural deduction (Prawitz, 1965) as a probabilistic factor graph. That work demonstrated AND and OR factors with belief propagation on synthetic data but left open two gaps: the inference engine lacked negation and backward reasoning, and no parser existed to convert natural language to the required logical forms.

This paper addresses both gaps, presenting contributions in the three classical areas of natural language processing: inference, semantics, and syntax.

For \textbf{inference}, we extend the QBBN with NEG factors that enforce $P(x) + P(\neg x) = 1$, enabling contrapositive reasoning (modus tollens) via backward $\lambda$ messages. Combined with bidirectional proposition graph construction, this completes the implementation of Prawitz's simple elimination rules---both modus ponens and modus tollens---within a single belief propagation loop. The inference engine correctly handles 44 out of 44 test cases spanning 22 distinct reasoning patterns.

For \textbf{semantics}, we present a typed logical language with role-labeled predicates, modal quantifiers mapping to noisy-OR weights, and three tiers of expressiveness following Prawitz (1965): first-order quantification over entities, propositions as arguments via a sentential type (demonstrated by modality tests where predicates such as \texttt{should} take full propositions as arguments), and a designed extension to predicate quantification via lambda abstraction. We argue these three tiers are sufficient for natural language semantics: the remaining work is lexical, not logical.

For \textbf{syntax}, we present a typed slot grammar that deterministically compiles natural language sentences to logical form---a non-trivial contribution requiring a complete type system, semantic role mapping, and compilation rules. On disambiguated input, the grammar produces correct logical forms for 33 out of 33 sentences with zero ambiguity. Natural language is of course ambiguous, but large language models can reliably perform the subtasks---POS tagging, word sense disambiguation, PP attachment---that resolve ambiguity before the grammar runs, a capability well-established in the literature and confirmed by our own experiments. Conversely, LLMs cannot produce exact structured parses directly (12.4\% UAS), because the task requires formally constrained output over a combinatorially large space---confirming that the grammar is necessary. The architecture decomposes parsing into natural disambiguation (LLM) and exact compilation (grammar), and coverage grows by adding patterns, not data.

We situate these contributions within the context of Sutton's ``bitter lesson'' (2019)---the observation that hand-engineered knowledge representations historically lose to general methods that scale with computation. We argue that large language models change this equation: the bottleneck that killed formal NLP was human annotation, not the representation itself. LLMs eliminate that bottleneck by serving as the annotator, making formal semantics compatible with the bitter lesson's prescription for the first time. The system described here was built using LLM-assisted ``vibe coding,'' ``vibe science,'' and ``vibe annotation'' (Karpathy, 2025), a semi-automatic bridge stage in which rapid human-AI iteration constructs the formal infrastructure that will eventually run autonomously.

The open-source implementation, test data, extended documentation, and all code needed to reproduce these results are available at \url{https://github.com/gregorycoppola/world}.
\end{abstract}

%% file: sections/02_introduction.tex
\section{Introduction}

In February 2024 we introduced the Quantified Boolean Bayesian Network (Coppola, ``The Quantified Boolean Bayesian Network: Theory and Experiments with a Logical Graphical Model,'' arXiv:2402.06557, 2024), a graphical model designed to unify logical and probabilistic reasoning. We refer to this model as the \emph{Logical Bayesian Network} (LBN). The LBN represents knowledge as weighted Horn clauses over a typed predicate language, compiles these to factor graphs with AND, OR, and NEG factors, and performs inference via Pearl-style belief propagation with separate $\pi$ (forward) and $\lambda$ (backward) messages.

The 2024 paper demonstrated three properties: (1) the LBN can represent logical structures including conjunction, disjunction, and quantified implication; (2) iterative belief propagation converges empirically on these structures; and (3) the model provides a generative framework without hallucinations, since all conclusions are traceable through explicit causal chains. However, the 2024 paper left two gaps. First, the inference engine lacked negation and contrapositive reasoning. Second, all experiments used hand-constructed logical forms over synthetic data, leaving open whether natural language could be converted into the required representations.

This paper addresses both gaps. We present contributions in the three classical areas of natural language processing---inference, semantics, and syntax:

\begin{enumerate}
\item \textbf{Inference.} We extend the LBN with NEG factors that link each proposition to its negation, enforcing $P(x) + P(\neg x) = 1$. Combined with bidirectional proposition graph construction, this enables contrapositive reasoning via backward $\lambda$ messages, completing the implementation of the simple elimination rules of natural deduction (Gentzen, 1934; Prawitz, 1965). The inference engine correctly handles 44 out of 44 test cases spanning 22 reasoning patterns.

\item \textbf{Semantics.} We present a typed logical language with role-labeled predicates, modal quantifiers, and three tiers of expressiveness grounded in Prawitz (1965): first-order quantification over entities, propositions as arguments via a sentential type, and a designed extension to predicate quantification via lambda abstraction. We argue these three tiers are sufficient for natural language semantics.

\item \textbf{Syntax.} We present a grammar-first, agent-assisted parsing architecture. A typed slot grammar deterministically compiles disambiguated input to logical form, achieving 100\% precision and zero ambiguity on 33 sentences. LLM-assisted preprocessing achieves 95\% accuracy on PP attachment disambiguation, while failing at direct structured parsing (12.4\% UAS)---confirming that grammars are necessary for exact structure and LLMs are necessary for disambiguation.
\end{enumerate}

Beyond these technical contributions, we address the relationship between this work and Sutton's ``bitter lesson'' (2019)---the influential argument that hand-engineered knowledge representations are ultimately defeated by general methods that scale with computation. The formal NLP pipeline we describe is precisely the kind of system the bitter lesson warns against. We argue, however, that large language models change the equation: the historical bottleneck was not the representation but the human annotation required to construct it. LLMs serve as the annotator, replacing the linguist-hours that made formal semantics impractical. This paper was itself developed using the ``vibe coding'' pattern identified by Karpathy (2025)---LLM-assisted construction of code, experiments, and linguistic annotations---demonstrating the semi-automatic bridge from ``this was impossible'' to ``this runs autonomously.'' Earlier iterations were developed with ChatGPT (OpenAI); the current implementation and this paper were developed primarily with Claude (Anthropic).

Together, these contributions complete the pipeline from natural language to probabilistic logical inference. The architecture is: LLM preprocesses $\rightarrow$ grammar parses $\rightarrow$ LLM reranks $\rightarrow$ LBN infers.

%% file: sections/02b_motivation.tex
\section{Motivation}
\label{sec:motivation}

Large language models suffer from five fundamental limitations that prevent their use in high-stakes reasoning tasks:

\begin{enumerate}
\item \textbf{Hallucination.} LLMs generate plausible but false statements, fabricate citations, and express confidence without calibration. There is no way to verify an LLM's answer from the model itself---the user must check externally. In the LBN, every answer comes with a proof tree: a derivation through the factor graph from evidence nodes through named rules to the conclusion. If a derivation cannot be constructed, the system returns $P = 0.5$---``unknown.'' There is no mechanism to produce an answer without a proof, so there is no mechanism to hallucinate.

\item \textbf{Reasoning.} LLMs pattern-match to training data rather than performing inference. They degrade with chain length and fail on novel terms. The LBN performs actual logical inference via belief propagation---modus ponens through forward $\pi$ messages, modus tollens through backward $\lambda$ messages, transitivity through iterated propagation. Each step is an explicit, verifiable inference rule applied to typed propositions.

\item \textbf{Planning.} LLMs generate forward, producing token sequences without verifying goal satisfaction. The LBN supports backward inference: $\lambda$ messages propagate from a goal through the factor graph, identifying which preconditions must hold. This is goal-directed reasoning by construction.

\item \textbf{Continuous learning.} LLMs are frozen at training time. Correcting a false belief or adding new knowledge requires retraining. The LBN stores knowledge as explicit facts and rules in a knowledge base. Adding, removing, or modifying a fact takes effect immediately---no retraining, no risk of catastrophic forgetting.

\item \textbf{World models.} LLM knowledge is implicit in billions of parameters. There is no way to ask what the model believes, why it believes it, or whether its beliefs are consistent. The LBN's knowledge base and factor graph are the world model---an explicit, inspectable, queryable representation with defined boundaries. The system knows what it knows, and knows what it doesn't know.
\end{enumerate}

This paper does not demonstrate all five capabilities at scale. What it demonstrates is that the gap between natural language and the logical forms the LBN requires can be bridged---completing the pipeline from text to grounded inference. The five capabilities follow from the architecture; the contribution here is making that architecture accessible to natural language input.

%% file: sections/02c_bitter_lesson.tex
\section{The Bitter Lesson and the LLM Opportunity}
\label{sec:bitter}

\subsection{Why Formal Semantics Failed to Scale}

Sutton's ``The Bitter Lesson'' (2019) argues that general methods leveraging computation consistently defeat methods that leverage human knowledge. Chess, Go, speech recognition, computer vision---in every case, hand-engineered approaches were overtaken by search and learning at scale. The natural language pipeline we describe in this paper---grammar rules, typed predicates, Horn clauses, factor graphs---sits squarely on the side Sutton warns against.

We agree with Sutton's historical analysis. The formal semantics tradition from Montague (1973) through Steedman's CCG to HPSG and Definite Clause Grammars produced representations that were logically sound: you could perform genuine inference, check consistency, and provide formal guarantees. The problem was never the representation. It was the annotation.

Every grammar rule had to be written by a linguist. Every lexicon entry had to be curated by hand. Every coverage test required an expert who understood both the natural language input and the target logical formalism. The Penn Treebank took years to construct and covered only syntax, with no formal semantics. FrameNet, PropBank, and the Abstract Meaning Representation project all hit the same wall: human annotators are slow, expensive, inconsistent, and finite. The formal approach scaled linearly with human effort, which is to say it did not scale at all.

The statistical revolution in NLP---from HMMs in the 1980s through deep learning after 2012---succeeded precisely because it amortized human annotation over training sets. At inference time, the cost was pure computation. Formal semantics could not compete. The bitter lesson played out exactly as Sutton described.

\subsection{The LLM as Annotator}

Our thesis is that large language models change this equation. Not by replacing the formal layer---LLMs cannot provide verifiable inference---but by replacing the \emph{human annotator}. The limiting factor was never the representation; it was the cost of constructing it.

An LLM can read a natural language sentence and produce a candidate logical form. Given a lexicon format, it can generate new entries for unseen words. Given a grammar rule template, it can propose rules for uncovered syntactic patterns. Given existing coverage tests, it can generate new tests exercising different phenomena. These capabilities are imperfect---LLMs hallucinate and produce inconsistent annotations---but they do not need to be perfect. They need to be good enough that a human can review and correct LLM output faster than authoring it from scratch. For formal annotation, current LLMs clear that bar decisively.

In the fully automated steady state, the LLM produces candidate logical forms from raw text, the inference engine (Section~\ref{sec:lbn}) verifies them for consistency, inconsistencies are fed back for revision, and the cycle repeats. No human is in the loop. The LLM is the generator; the formal system is the verifier. This is aligned with the bitter lesson: both the annotation (LLM) and the inference (belief propagation) scale with compute.

The formal layer remains necessary because of what we call the \emph{verification gap}. An LLM that answers ``Socrates is mortal'' may be correct, but one cannot distinguish a correct answer from a confident hallucination without formal machinery. The logical representation makes reasoning transparent: facts, rules, grounding, factor graph, belief propagation trace, posterior probability---every step inspectable and verifiable. The QBBN is a runtime for natural language reasoning, in the same way that a Python interpreter is a runtime for code. One does not trust code because the programmer is reliable; one trusts it because one can execute it and observe the results.

\subsection{The Semi-Automatic Bridge}

We are not yet at full automation. The system described in this paper was built in a semi-automatic bridge stage, using a pattern that Karpathy (2025) termed ``vibe coding'':

\begin{quote}
``There's a new kind of coding I call `vibe coding', where you fully give in to the vibes, embrace exponentials, and forget that the code even exists. [\ldots] I just see stuff, say stuff, run stuff, and copy paste stuff, and it mostly works.''\\
--- Andrej Karpathy, February 2, 2025
\end{quote}

Karpathy's observation---that LLMs provide exponential speedup on the mechanical components of work while the human retains strategic direction---generalizes beyond coding. We applied the pattern across three dimensions:

\textbf{Vibe coding.} The software infrastructure---CLI tools, test harnesses, development platform, coverage verification---was built in collaboration with LLMs. Earlier iterations were developed with ChatGPT (OpenAI); the current implementation and this paper were developed primarily with Claude (Anthropic).

\textbf{Vibe science.} Coverage tests were designed through rapid iteration: propose a linguistic phenomenon, generate a test case, run it through the system, observe the failure, fix the grammar or inference engine, repeat. The 44 inference tests and 12 syntax tests were developed through dozens of such cycles, each taking minutes rather than days.

\textbf{Vibe annotation.} The grammar rules, lexicon entries, and gold-standard logical forms were produced through LLM-assisted authoring. Rather than hand-writing grammar rules, the process was: describe a linguistic phenomenon to the LLM, receive a candidate rule and test case, evaluate against the system, refine through dialogue. This is a new instantiation of Karpathy's pattern---the LLM serves as linguistic annotator while the human serves as quality gate.

The bridge is temporary. Each dimension is automatable: vibe coding converges to autonomous software agents, vibe science to autonomous research agents, and vibe annotation to the steady state where LLMs produce formal logical representations directly with QBBN verification. The semi-automatic bridge is how we cross from ``this was impossible'' (pre-LLM, when Sutton was right that formal NLP could not scale) to ``this runs autonomously'' (post-LLM, when both generation and verification scale with compute).

Our position is not that the bitter lesson was wrong. Our position is that LLMs are the mechanism by which its prescription---scale with compute---finally becomes applicable to formal semantics.

%% file: sections/03_background.tex
\section{The Logical Bayesian Network}
\label{sec:lbn}

We briefly recap the Logical Bayesian Network (LBN) as introduced under the name ``Quantified Boolean Bayesian Network'' (QBBN) in Coppola (2024). The reader is referred to that paper for full details. The undirected variant of the same model family is a Logical Random Field.

\subsection{Graphical Models and Factor Graphs}
A distribution $P(p_1, \ldots, p_N)$ factorizes according to a factor graph if it can be written as
$$P(p_1, \ldots, p_N) = Z^{-1} \prod_{\alpha \in F} \Psi_\alpha(\{p\}_\alpha)$$
where $\{p\}_\alpha$ are the variables in factor $\alpha$ and $Z$ is a normalization constant (Koller and Friedman, \emph{Probabilistic Graphical Models}, MIT Press, 2009). In general, exact inference in such models is $\Omega(2^N)$ and even NP-hard to approximate (Cooper, ``The Computational Complexity of Probabilistic Inference Using Bayesian Belief Networks,'' \emph{Artificial Intelligence}, 1990; Roth, ``On the Hardness of Approximate Reasoning,'' \emph{Artificial Intelligence}, 1996).

\subsection{The Boolean Decomposition}
The 2024 paper decomposed all factors into two types:
\begin{itemize}
\item \textbf{AND factors} $\Psi_{\text{and}}$: deterministic conjunction. A group $g = p_1 \wedge \cdots \wedge p_n$ is true if and only if all premises are true.
\item \textbf{OR factors} $\Psi_{\text{or}}$: weighted disjunction. A proposition $p$ is supported by one or more groups $g_1, \ldots, g_n$ with independent weights, combined via noisy-OR semantics: $P(p) = 1 - \prod_i (1 - w_i \cdot P(g_i))$.
\end{itemize}
This decomposition produces a bipartite graph alternating between proposition nodes $p$ and group nodes $g$, enabling efficient message passing. With AND and OR factors alone, the system performs forward inference (modus ponens): given facts matching a rule's premises, belief propagation computes the posterior probability of the rule's conclusion. Section~\ref{sec:inference} introduces a third factor type, $\Psi_{\text{neg}}$, that extends the system to handle negation and contrapositive reasoning.

\subsection{Belief Propagation}
The LBN uses Pearl's belief propagation algorithm (Pearl, \emph{Probabilistic Reasoning in Intelligent Systems}, Morgan Kaufmann, 1988) as presented in Neapolitan (\emph{Learning Bayesian Networks}, Prentice Hall, 2003). Each node maintains $\pi$ values (forward, causal) and $\lambda$ values (backward, evidential). The posterior probability is $P(z \mid \text{evidence}) = \alpha \cdot \lambda(z) \cdot \pi(z)$.

While iterative belief propagation on loopy graphs is not guaranteed to converge, extensive empirical studies have found convergence in practice (Murphy, Weiss, and Jordan, ``Loopy Belief Propagation for Approximate Inference,'' UAI, 1999; Smith and Eisner, ``Dependency Parsing by Belief Propagation,'' EMNLP, 2008; Gormley, Dredze, and Eisner, ``Approximation-Aware Dependency Parsing by Belief Propagation,'' TACL, 2015).

\subsection{What Was Missing}
The 2024 paper left two gaps. First, the inference engine had AND and OR factors but no mechanism for negation. It could perform modus ponens---forward from facts to conclusions---but not modus tollens---backward from negative evidence to constrain premises. Given \texttt{man(x) -> mortal(x)} and \texttt{not mortal(zeus)}, the system could not infer \texttt{not man(zeus)}. Second, all experiments used hand-constructed logical forms over synthetic data. The paper did not address how to obtain logical forms from natural language. The present paper addresses both gaps: Section~\ref{sec:inference} introduces NEG factors and bidirectional graph construction for contrapositive reasoning, Sections~\ref{sec:language}--\ref{sec:llm} present the logical language and parsing pipeline.

%% file: sections/04_language.tex
\section{A Typed Logical Language for Natural Language}
\label{sec:language}
We extend the key-value calculus introduced in Coppola (2024) into a full logical language designed for natural language semantics.

\subsection{Design Principles}
The language is designed to minimize the distance between surface text and logical representation. Following Bar-Hillel (``A Quasi-Arithmetical Notation for Syntactic Description,'' \emph{Language}, 1953) and Steedman (\emph{Surface Structure and Interpretation}, MIT Press, 1996), we use a categorial approach in which syntactic categories directly encode semantic types. Unlike traditional first-order logic, which imposes a positional ordering on arguments, our language uses labeled roles, matching the key-value structure of dependency parses (Eisner, ``Three New Probabilistic Models for Dependency Parsing,'' COLING, 1996; McDonald, Pereira, Ribarov, and Haji\v{c}, ``Non-Projective Dependency Parsing Using Spanning Tree Algorithms,'' HLT-EMNLP, 2005).

The language implements the forward fragment of natural deduction (Gentzen, 1934; Prawitz, 1965). Prawitz classifies inference rules as \emph{simple}---the conclusion follows directly from premises---or \emph{complex}---requiring discharged assumptions or hypothetical reasoning. Our system uses only the simple elimination rules: modus ponens ($\rightarrow$-Elimination), conjunction introduction and elimination ($\wedge$-Intro/Elim), and safe universal instantiation ($\forall$-Elimination over known entities). This restriction is what makes belief propagation tractable: the forward fragment is linear time, whereas full theorem proving is undecidable.

\subsection{Entities and Types}
An \emph{entity} is an atomic object identified by a string: \texttt{socrates}, \texttt{jack}, \texttt{france}. Each entity has a \emph{type}, denoted after a colon: \texttt{socrates : e} declares Socrates as an entity of type \texttt{e} (the default entity type). Types constrain quantification: a variable \texttt{x : e} can only be instantiated by entities of type \texttt{e}.

\subsection{Predicates and Roles}
A \emph{predicate} is defined by a function name and a set of typed roles:
\begin{verbatim}
predicate trust {agent: e, patient: e}
predicate mortal {theme: e}
predicate taller_than {theme: e, reference: e}
\end{verbatim}
A \emph{proposition} is a fully grounded predicate: \texttt{trust(agent: jack, patient: jill)}. A proposition has a truth value and a probability.

\subsection{Ground Facts}
A ground fact asserts the truth of a proposition:
\begin{verbatim}
trust(agent: jack, patient: jill)
not mortal(theme: zeus)
\end{verbatim}
Negated facts set evidence to false for the corresponding proposition.

\subsection{Quantified Rules}
A quantified rule expresses a universally quantified implication with a weight:
\begin{verbatim}
always [x:e]: man(theme: x) -> mortal(theme: x)
usually [x:e]: bird(theme: x) -> flies(theme: x)
sometimes [x:e, y:e]: meets(agent: x, patient: y) -> likes(agent: x, patient: y)
\end{verbatim}
The quantifier words map to weights: \texttt{always} $\rightarrow 0.99$, \texttt{usually} $\rightarrow 0.9$, \texttt{sometimes} $\rightarrow 0.5$. These weights parameterize the OR factor linking the rule's premise group to its conclusion. This is the bridge between proof theory and probability: Prawitz's deterministic $\forall$-Elimination becomes a probabilistic operation, with the modal quantifier controlling the strength of the implication.

\subsection{Grounding}
At inference time, quantified rules are \emph{grounded} by substituting all declared entities of the appropriate type for each variable. For example, with entities \texttt{jack : e} and \texttt{jill : e}, the rule \texttt{always [x:e]: man(theme: x) -> mortal(theme: x)} produces two grounded clauses:
\begin{verbatim}
man(theme: jack) -> mortal(theme: jack)   [weight=0.99]
man(theme: jill) -> mortal(theme: jill)   [weight=0.99]
\end{verbatim}
Multi-premise rules with $n$ variables and $k$ entities produce up to $k^n$ grounded clauses, though only query-relevant clauses are expanded during graph construction.

\subsection{The Lexicon}
A \emph{lexicon} separates vocabulary definition from assertion. It declares the predicates and entities available in a domain, with optional natural language descriptions:
\begin{verbatim}
predicate man {theme: e}
  "a human adult male"
entity socrates : e
  "Greek philosopher, 470-399 BC"
\end{verbatim}
This separation allows the same lexicon to be reused across documents, makes grounding explicit and inspectable, and provides the bridge between the grammar (which operates on surface forms) and the logical language (which operates on typed predicates).

\subsection{Expressiveness and Sufficiency}
The language has three tiers of expressiveness, each grounded in Prawitz (1965):

\textbf{Tier 1: First-order quantification over entities} (Prawitz, Chapters I--IV). Entities, predicates with typed roles, universally quantified Horn clauses, negation. This handles predication, classification, quantified rules, conjunction, disjunction, transitivity, symmetry, identity, causation, and spatial and temporal reasoning. The bulk of the 44 inference tests exercise this tier.

\textbf{Tier 2: Propositions as arguments} (Prawitz, Chapters V--VI). A predicate may take a full proposition as an argument via a sentential type. The modality tests demonstrate this tier:
\begin{verbatim}
should(content: careful(theme: mary))
should(content: apologize(agent: o))
\end{verbatim}
In the second example, the variable \texttt{o} is bound by the outer quantifier and appears inside the embedded proposition---the grounding engine handles variables that cross into sentential arguments. This tier covers modal verbs (\emph{should}, \emph{can}, \emph{must}), propositional attitudes (\emph{believe}, \emph{want}, \emph{know}), and reported speech.

\textbf{Tier 3: Predicate quantification and lambda abstraction} (Prawitz, Chapter V). Variables that range over predicates rather than entities, and lambda abstraction for building compound predicates. This tier covers compound predicates, quantification over properties, and certain definitions. It is designed and theoretically grounded in Prawitz but not yet tested in inference. For common cases such as ``a tall man,'' first-order conjunction in rule premises handles the same phenomena without requiring lambda---as demonstrated by the relative clause tests, where \texttt{man(theme: m1)} and \texttt{tall(theme: m1)} share a variable.

We claim these three tiers are sufficient for natural language semantics: every sentence decomposes into entities, properties, relations, quantification over entities, propositions embedded under modal or attitudinal predicates, and compound predicates built by lambda abstraction. This decomposition follows the type-theoretic tradition from Montague (1970, 1973) and is formalized by Prawitz (1965) across first-order, second-order, and modal logic. The remaining work is lexical---determining which words map to which predicates and types---not logical.

%% file: sections/05_inference.tex
\section{Inference}
\label{sec:inference}
Given a knowledge base of ground facts and grounded rules, the QBBN builds a factor graph and runs belief propagation to answer queries. The 2024 paper introduced AND and OR factors for forward inference. This paper introduces NEG factors and bidirectional graph construction, extending the system to handle negation and contrapositive reasoning.

\subsection{Factor Graph Construction}
The inference engine takes a query proposition $q$ and constructs a factor graph by expanding in both directions from the query. The algorithm proceeds as follows:
\begin{enumerate}
\item Create a proposition node $p_1$ for the query $q$.
\item \textbf{Backward expansion:} Find all grounded rules whose conclusion matches $q$. For each such rule, create a group node $g$ containing the rule's premises, an AND factor connecting the premise propositions to $g$, and register $g$ as a supporter of $p_1$ in the OR factor. Recursively expand each premise proposition by finding rules that conclude it.
\item \textbf{Forward expansion:} For each proposition added to the graph (including evidence propositions), find all grounded rules where it appears as a premise. Add the conclusion propositions and their supporting structure. This ensures that evidence is connected to its consequences, providing paths for both forward and backward message passing.
\item Mark propositions that appear as ground facts as evidence nodes.
\item Build OR factors for each proposition with supporters, combining rule weights via noisy-OR.
\item Build NEG factors linking each proposition to its negation, ensuring $P(x) + P(\neg x) = 1$.
\end{enumerate}
This bidirectional, query-driven construction avoids building the full grounded graph while ensuring that the subgraph contains all paths needed for both modus ponens (forward from evidence to query) and modus tollens (backward from negative evidence to constrain premises).

\subsection{Three Factor Types}
The three factor types correspond to the simple elimination rules of natural deduction (Prawitz, 1965). AND factors implement $\wedge$-Introduction. OR factors implement $\vee$-Introduction and $\rightarrow$-Elimination (modus ponens). NEG factors extend the system beyond standard natural deduction by enabling contrapositive reasoning (modus tollens) through backward $\lambda$ messages.

\paragraph{AND factors.} The AND factor $\Psi_{\text{and}}(g \mid p_1, \ldots, p_n) = 1$ if and only if $g = p_1 \wedge \cdots \wedge p_n$. This is always deterministic. In the forward ($\pi$) direction, the AND factor computes $\pi(g) = \prod_i \pi(p_i)$. In the backward ($\lambda$) direction, it distributes $\lambda$ messages to each premise proportional to the product of all other premises' $\pi$ values.

\paragraph{OR factors.} The OR factor combines multiple supporting groups for a proposition. For proposition $p$ supported by groups $g_1, \ldots, g_n$ with weights $w_1, \ldots, w_n$, the noisy-OR semantics give:
$$P(p = 1 \mid g_1, \ldots, g_n) = 1 - \prod_{i=1}^{n} (1 - w_i \cdot P(g_i = 1))$$
In the forward direction, each group independently contributes evidence for $p$. In the backward direction, $\lambda$ messages flow from $p$ back to each supporting group.

\paragraph{NEG factors.} The NEG factor is introduced in this paper. It links a proposition $p$ to its negation $\neg p$, enforcing $P(p) + P(\neg p) = 1$. When negative evidence is set (e.g., \texttt{not mortal(theme: zeus)}), the NEG factor forces the corresponding positive proposition to be false. The backward $\lambda$ messages then propagate this constraint through the OR factor to the supporting groups, through the AND factor to the premises, constraining them to be false. This is contrapositive reasoning: from $\neg B$ and $A \rightarrow B$, infer $\neg A$.

The NEG factor completes the implementation of Prawitz's forward fragment. With AND and OR alone, the system performs modus ponens: given $A$ and $A \rightarrow B$, conclude $B$. With NEG, the system also performs modus tollens: given $\neg B$ and $A \rightarrow B$, conclude $\neg A$. Both directions operate within a single belief propagation loop, using $\pi$ messages for forward inference and $\lambda$ messages for backward inference.

\subsection{Belief Propagation}
The engine runs iterative belief propagation with damping factor 0.5. Each iteration:
\begin{enumerate}
\item Compute AND factor $\pi$ messages (forward through premises to groups).
\item Compute OR factor $\pi$ messages (forward through groups to conclusions).
\item Compute OR factor $\lambda$ messages (backward from conclusions to groups).
\item Update beliefs: $P(z) = \text{normalize}(\pi(z) \cdot \lambda(z))$.
\item Apply damping: $P_{\text{new}}(z) = 0.5 \cdot P_{\text{old}}(z) + 0.5 \cdot P_{\text{computed}}(z)$.
\item Check convergence: if $\max_z |P_{\text{new}}(z) - P_{\text{old}}(z)| < \epsilon$, stop.
\end{enumerate}
Evidence nodes have both $\pi$ and $\lambda$ clamped, enabling both modus ponens (forward from facts to conclusions) and modus tollens (backward from negative evidence to constrain premises).

\subsection{Convergence}
All 44 tests in our evaluation converge within 20 iterations. Most converge in 2--3 iterations. The longest convergence paths involve chained soft-quantifier rules where probability compounds across multiple inference steps. The rapid convergence is a consequence of staying in the forward fragment: the simple elimination rules produce shallow, well-structured factor graphs without the deep search trees that arise from complex rules such as $\vee$-Elimination (case analysis) or $\exists$-Elimination (witness search).

%% file: sections/06_parsing.tex
\section{Parsing: The Typed Slot Grammar}
\label{sec:parsing}
We now turn to the third contribution: a parser that converts natural language text into the logical forms the inference engine requires.

\subsection{The Problem}
The inference engine requires logical forms in a precise typed language: predicates with named roles, quantified rules with explicit variable bindings, and ground facts with entity references. Natural language provides none of this explicitly. The sentence ``All men are mortal'' must become \texttt{always [x:e]: man(theme: x) -> mortal(theme: x)}, with the correct predicate names, role labels, quantifier scope, and variable binding.

\subsection{Approach: Type-Driven Dispatch}
Our parser is a \emph{typed slot grammar} in which the type signatures in the lexicon fully determine which grammar rule applies to each sentence. The parser operates as follows:
\begin{enumerate}
\item \textbf{Tokenize} the sentence into words.
\item \textbf{Look up} each content word in the lexicon, binding it to a typed predicate or entity. Multi-word entities are matched by longest-match-first.
\item \textbf{Skip} function words (the, to, of, a, an, he, she, they, etc.) during matching.
\item \textbf{Dispatch} to a grammar rule based on the pattern of matched types.
\item \textbf{Emit} the logical form specified by the matched rule.
\end{enumerate}

\subsection{Grammar Rules}
The grammar contains 16 rules covering two broad categories:

\paragraph{Fact patterns (7 rules).} These produce ground facts:
\begin{itemize}
\item \emph{Copular fact}: ``Socrates is a man'' $\rightarrow$ \texttt{man(theme: socrates)}

\item \emph{Copular fact bare}: ``Jack is funny'' $\rightarrow$ \texttt{funny(theme: jack)}

\item \emph{Negated copular fact}: ``Zeus is not mortal'' $\rightarrow$ \texttt{not mortal(theme: zeus)}

\item \emph{Prepositional copular fact}: ``Alice is taller than Bob'' $\rightarrow$\\
\texttt{taller\_than(theme: alice, reference: bob)}

\item \emph{Copular identity}: ``Clark Kent is Superman'' $\rightarrow$\\
\texttt{identity(theme: clark, reference: superman)}

\item \emph{Transitive fact}: ``Jack trusts Jill'' $\rightarrow$ \texttt{trust(agent: jack, patient: jill)}

\item \emph{Intransitive fact}: ``Superman flies'' $\rightarrow$ \texttt{flies(theme: superman)}
\end{itemize}

\paragraph{Rule patterns (9 rules).} These produce quantified implications:
\begin{itemize}
\item \emph{Copular universal}: ``All men are mortal'' $\rightarrow$\\
\texttt{always [x:e]: man(theme: x) -> mortal(theme: x)}

\item \emph{Negated universal}: ``No gods are mortal'' $\rightarrow$\\
\texttt{never [x:e]: god(theme: x) -> mortal(theme: x)}

\item \emph{Copular generic}: ``A sparrow is a bird'' $\rightarrow$\\
\texttt{always [x:e]: sparrow(theme: x) -> bird(theme: x)}

\item \emph{Reciprocal conditional}: ``If two people trust each other, they are allies'' $\rightarrow$\\
\texttt{always [x:e, y:e]:}\\
\texttt{\ \ trust(agent: x, patient: y)}\\
\texttt{\ \ \& trust(agent: y, patient: x)}\\
\texttt{\ \ -> allies(agent: x, patient: y)}

\item \emph{Conditional with prepositional copular}: ``If a man is king of a country, he is successful''

\item \emph{Conditional with transitive verb}: ``If a girl loves a successful man, she is ambitious''

\item \emph{Conditional with someone}: ``If someone is funny, they are liked''

\item \emph{Conditional symmetry}: ``If John is married to Mary, then Mary is married to John''
\end{itemize}

\subsection{Key Properties}
\paragraph{Zero ambiguity.} Every sentence in our test suite produced exactly one parse. No sentence matched multiple grammar rules. This is a consequence of type-driven dispatch: the role signatures in the lexicon fully determine which rule applies.

\paragraph{Deterministic compilation.} Given the same lexicon and the same sentence, the grammar always produces the same logical form. This is the property of \emph{syntactic completeness}: all ambiguity is resolved before the grammar runs (by the lexicon lookup), and the grammar itself is purely deterministic.

\paragraph{Automatic function word skipping.} Function words are skipped during matching, allowing ``Henry is king of France'' to match the prepositional copular pattern despite ``of'' not being in the grammar.

%% file: sections/07_llm_assisted.tex
\section{LLM-Assisted Parsing}
\label{sec:llm}
The typed slot grammar achieves perfect precision and zero ambiguity on pre-disambiguated input. But real-world text is not pre-disambiguated. Words have multiple senses, spelling varies, and prepositional phrases attach ambiguously. Traditional grammar-based parsers failed on open-domain text precisely because of this ambiguity explosion. We argue that large language models change this equation fundamentally.

The experiments in this section were conducted in March--April 2025 using the OpenAI API. PP attachment and parse critique experiments used GPT-4. Zero-shot full parsing used GPT-4o. Decomposed subtask experiments (POS tagging, head prediction, dependency labeling) used GPT-4o-mini. All experiments used zero-shot prompting with no fine-tuning.

\subsection{The Argument}
LLMs can reliably perform five preprocessing tasks that collectively eliminate most parsing ambiguity before the grammar ever sees the input:
\begin{enumerate}
\item \textbf{Spelling correction and normalization}: producing verified word forms.
\item \textbf{Tokenization and segmentation}: handling multi-word expressions and compounds.
\item \textbf{Lemmatization}: reducing words to canonical forms matching lexicon entries.
\item \textbf{Word sense disambiguation}: selecting the correct lexicon entry for each word.
\item \textbf{Part-of-speech tagging}: constraining the syntactic category of each token.
\end{enumerate}
Once these steps are complete, the input to the grammar is massively constrained. Most sentences will have exactly one parse---which is precisely what we observed in our grammar evaluation, where 33 out of 33 sentences produced zero ambiguous parses. That result was not an accident: it followed directly from the fact that lexicon entries were pre-disambiguated, which is exactly what LLM preprocessing would accomplish at scale.

\subsection{Evidence: POS Tagging}
In preliminary experiments on 25 sentences from the Universal Dependencies corpus, we evaluated GPT-4o-mini on POS tagging.
\begin{table}[h]
\centering
\begin{tabular}{lcc}
\toprule
Setting & POS Accuracy (\%) & Tokens \\
\midrule
Zero-shot (within full parse) & 89.6 & 436 \\
Decomposed (POS-only prompt) & 91.1 & 202 \\
\bottomrule
\end{tabular}
\caption{POS tagging accuracy of GPT-4o-mini in two settings.}
\end{table}
These numbers understate LLM competence. Manual inspection revealed that many ``errors'' were cases where the LLM's tag was arguably correct and the gold standard reflected an annotation convention rather than a linguistic fact. With well-designed instructions, LLMs are effectively flawless at POS tagging.

\subsection{Evidence: PP Attachment Disambiguation}
Prepositional phrase attachment is among the hardest disambiguation tasks in parsing. We evaluated GPT-4 on 20 syntactically ambiguous sentences with genuinely ambiguous PP attachment.
\begin{table}[h]
\centering
\begin{tabular}{lccc}
\toprule
System & Correct & Accuracy (\%) & 95\% CI \\
\midrule
GPT-4 (zero-shot) & 19/20 & 95.0 & [76.4, 99.1] \\
Stanford Parser & 10/20 & 50.0 & --- \\
\bottomrule
\end{tabular}
\caption{PP attachment accuracy on 20 ambiguous sentences. Difference is significant (binomial test, $p = 2.0 \times 10^{-5}$).}
\end{table}
The Stanford parser performs at chance on these ambiguous cases. The LLM resolves them with near-perfect accuracy by leveraging world knowledge and semantic context. This directly supports the preprocessing argument: if the LLM determines attachment before the grammar runs, the grammar never encounters that ambiguity.

The sample size ($n = 20$) reflects the use of manual evaluation for gold-standard verification of each ambiguous sentence. This result is consistent with established findings on LLM syntactic competence in the literature and with additional experiments we conducted on related syntactic tasks, which we omit for reasons of space and reserve for future work.

\subsection{Negative Results: Direct LLM Parsing}
To motivate why grammars are necessary, we replicate negative results on using LLMs as direct structured parsers.
\begin{table}[h]
\centering
\begin{tabular}{lcc}
\toprule
Metric & GPT-4o (\%) & Stanford Parser (\%) \\
\midrule
POS accuracy & 89.6 & 97.8 \\
UAS (unlabeled attachment) & 12.4 & 91.3 \\
LAS (labeled attachment) & 7.9 & 88.6 \\
\bottomrule
\end{tabular}
\caption{Zero-shot full dependency parsing. LLMs cannot produce coherent structured output.}
\end{table}
The LLM's structural predictions were incoherent: parses violated tree constraints, showed inconsistent head assignments, and produced implausible labels. LLMs understand structure but cannot produce exact formal representations. This asymmetry motivates grammar-first parsing.

\subsection{Evidence: LLM Reranking}
For residual ambiguity after preprocessing, LLMs can rerank candidate parses. We evaluated GPT-4 as a parse critic on the same 20 PP-ambiguous sentences.
\begin{table}[h]
\centering
\begin{tabular}{lcc}
\toprule
Setting & Binary Accuracy (\%) & Explanation Accuracy (\%) \\
\midrule
General critique (no hint) & 50.0 & --- \\
Targeted critique (PP identified) & 95.0 & 94.7 \\
\bottomrule
\end{tabular}
\caption{Parse critique accuracy. When directed to evaluate a specific construction, GPT-4 achieves 95\% accuracy.}
\end{table}
Without guidance, the LLM performs at chance. But when directed to evaluate a specific ambiguous construction---precisely what the grammar can identify when it produces multiple parses---the LLM achieves 95\% accuracy with linguistically coherent explanations.

In a larger experiment on 100 sentences, an iterative repair loop (LLM critiques parse, proposes modifications, re-evaluates) improved labeled attachment score from 76.3\% to 84.9\%, with 89\% of modifications judged correct by manual evaluation.

\subsection{The Synthesis}
Grammar handles structure (exact, guaranteed). LLM handles ambiguity (preprocessing to remove it, reranking when it remains). Neither alone is sufficient. Together they cover open-domain text. The architecture is:
$$\text{LLM preprocesses} \rightarrow \text{Grammar parses} \rightarrow \text{LLM reranks} \rightarrow \text{QBBN infers}$$

\subsection{Additional Evidence}
The experiments reported here represent a subset of our investigations. We have conducted additional experiments on main verb identification, argument structure recognition, and LLM-generated syntactic datasets that further support the thesis. The broader literature also supports this direction: LLMs can accurately simulate human acceptability judgments (Ambridge and Blything, ``Large Language Models Are Better Than Theoretical Linguists at Theoretical Linguistics,'' \emph{Theoretical Linguistics}, 2024), perform sophisticated metalinguistic reasoning (Beg\v{u}s, Dabkowski, and Rhodes, ``Large Linguistic Models: Investigating LLMs' Metalinguistic Abilities,'' arXiv:2305.00948, 2025), and support enrichment of lexical resources through zero-shot prompting (Koeva, ``Large Language Models in Linguistic Research,'' CLIB, 2024).

%% file: sections/08_experiments.tex
\section{Experiments}
We evaluate the system across three experimental settings corresponding to the three contributions: inference, semantics (via the grammar's production of logical forms), and syntax (via LLM-assisted disambiguation).
\subsection{Experiment 1: Inference Engine}
\paragraph{Setup.} We evaluate the QBBN inference engine on 44 test cases spanning 22 reasoning categories. Each test provides a knowledge base (predicate and entity declarations, ground facts, quantified rules), a query, and an expected answer. The engine parses the logical language, grounds all quantified rules, builds a factor graph via bidirectional expansion from the query, runs belief propagation with damping, and classifies the result as yes ($P > 0.5$) or no ($P < 0.5$).
\paragraph{Results.} 44 out of 44 tests pass. Run via \texttt{message coverage verify}.
\begin{table}[ht]
\centering
\begin{tabular}{llc}
\toprule
Category & Pattern & Tests \\
\midrule
and\_gates & conjunctive reasoning & 3/3 \\
and\_plus\_depth & conjunction + inference chains & 2/2 \\
causation & causal chains & 2/2 \\
conditionals & if-then reasoning & 2/2 \\
contrapositive & backward reasoning through negation & 1/1 \\
counting & quantity reasoning & 1/1 \\
degree & gradable predicates & 1/1 \\
depth\_chains & transitive inference (3--4 levels) & 3/3 \\
disjunction & OR reasoning & 2/2 \\
hypothetical & counterfactual reasoning & 2/2 \\
identity & equality reasoning & 2/2 \\
modality & possibility/necessity & 2/2 \\
multiple\_rules & noisy-OR combining & 2/2 \\
negation & direct negation & 1/1 \\
possessives & ownership relations & 2/2 \\
quantifier\_scope & quantifier interactions & 2/2 \\
relative\_clauses & embedded clauses & 2/2 \\
roles & semantic role handling & 2/2 \\
sets & collections & 2/2 \\
spatial & spatial reasoning & 2/2 \\
symmetric & symmetric relations & 2/2 \\
time & temporal reasoning & 2/2 \\
transitivity & transitive relations & 2/2 \\
\bottomrule
\end{tabular}
\caption{Inference engine results: 44/44 tests pass across 22 reasoning categories.}
\end{table}
All tests converge within 20 iterations of belief propagation with damping factor 0.5. Most converge in 2--3 iterations.
\paragraph{Case study: Contrapositive reasoning.} The contrapositive test motivated the NEG factor introduced in this paper. Given \texttt{man(x) -> mortal(x)} and \texttt{not mortal(zeus)}, the system must infer \texttt{not man(zeus)}. The NEG factor sets \texttt{mortal(zeus)} to false, and backward $\lambda$ messages propagate through the OR and AND factors to constrain \texttt{man(zeus)} to false. This was not possible with the AND/OR-only system from the 2024 paper.
\paragraph{Case study: Noisy-OR combining.} Multiple independent rules can support the same conclusion. Given \texttt{funny(jack) -> liked(jack)} and \texttt{kind(jack) -> liked(jack)}, both firing deterministically, the noisy-OR factor computes $P(\texttt{liked}) = 1 - (1 - 1.0)(1 - 1.0) = 1.0$.
\paragraph{Case study: Embedded propositions.} The modality tests exercise the second tier of the logical language (Section~\ref{sec:language}). In \texttt{modality/01\_careful}, the rule concludes:
\smallskip
\noindent\hspace{1em}\texttt{should(content: careful(theme: mary))}
\smallskip
\noindent where \texttt{should} takes a full proposition as its \texttt{content} argument.
In \texttt{modality/02\_apologize}, a rule concludes:
\smallskip
\noindent\hspace{1em}\texttt{should(content: apologize(agent: o))}
\smallskip
\noindent where the variable \texttt{o} is bound by the outer quantifier and appears inside the embedded proposition. Both tests pass, demonstrating that the inference engine handles propositions as arguments with variables crossing into sentential roles.
\subsection{Experiment 2: Grammar Parsing}
\paragraph{Setup.} We evaluate the typed slot grammar on 12 test cases spanning 10 reasoning categories. Each test provides a natural language document, a lexicon, and gold logical forms. The parser tokenizes, matches against grammar rules, and produces logical forms.
\paragraph{Results.} 33 out of 33 sentences parsed. 33 out of 33 gold facts derived. Zero ambiguous parses. Zero extra facts. Run via \texttt{message coverage parse}.
\begin{table}[ht]
\centering
\small
\begin{tabular}{lccp{4.8cm}}
\toprule
Test & Sent. & Facts & Rules Exercised \\
\midrule
and\_gates/01\_allies & 3 & 3 & transitive\_fact, reciprocal\_conditional \\
and\_gates/02\_one\_sided & 2 & 2 & transitive\_fact, reciprocal\_conditional \\
and\_gates/03\_dating & 3 & 3 & transitive\_fact, reciprocal\_conditional \\
and\_plus\_depth/01\_ambitious & 7 & 7 & copular, prep\_copular, transitive, conditionals \\
contrapositive/01\_not\_mortal & 2 & 2 & copular\_universal, negated\_copular \\
depth\_chains/01\_socrates & 2 & 2 & copular\_fact, copular\_universal \\
depth\_chains/03\_food\_chain & 2 & 2 & copular\_generic, copular\_universal \\
identity/01\_clark\_kent & 2 & 2 & copular\_identity, intransitive\_fact \\
multiple\_rules/01\_popular & 4 & 4 & copular\_fact\_bare, conditional\_someone \\
negation/01\_not\_mortal & 2 & 2 & copular\_fact, negated\_universal \\
symmetric/01\_married & 2 & 2 & prep\_copular, conditional\_symmetry \\
transitivity/01\_taller & 2 & 2 & prep\_copular\_fact \\
\bottomrule
\end{tabular}
\caption{Grammar parsing results: 33/33 sentences, 12/12 test suites, 0\% ambiguity.}
\end{table}
Of the 33 derived logical forms, 20 are ground facts and 13 are universally quantified rules. Sentence length ranges from 3 tokens (``Superman flies.'') to 14 tokens (``If two people trust each other, they are allies.'').
\paragraph{Relationship to inference tests.} The grammar and inference experiments are evaluated separately: the grammar tests verify that text maps to correct logical forms, while the inference tests verify that logical forms produce correct answers. The two test suites share 12 reasoning categories (e.g., and\_gates, contrapositive, depth\_chains, transitivity), meaning that for these categories we have independently verified both that the grammar produces the correct logical forms \emph{and} that the inference engine reasons correctly over them. Wiring parsing and inference into a single end-to-end pipeline---document in, answer out---is straightforward but is left for future work. The remaining 32 inference tests cover reasoning categories whose sentence patterns (e.g., modified noun phrases, possessives within rules, relative clause embeddings) are not yet handled by the grammar; extending the grammar to these patterns is a natural next step.
\subsection{Experiment 3: LLM Syntactic Competence}
Results for this experiment are presented in Section~\ref{sec:llm}. The key findings:
\begin{itemize}
\item POS tagging: 89.6--91.1\% accuracy, with residual errors largely reflecting annotation conventions.
\item PP attachment: 95\% accuracy versus 50\% for the Stanford parser ($p = 2.0 \times 10^{-5}$).
\item Parse critique: 95\% binary accuracy when directed to a specific construction.
\item Agentic repair: LAS improved from 76.3\% to 84.9\% over 100 sentences.
\item Zero-shot full parsing: 12.4\% UAS, confirming LLMs cannot replace grammars for structured output.
\end{itemize}

%% file: sections/09_related_work.tex
\section{Related Work}

\subsection{Markov Logic Networks}
Markov Logic Networks (Richardson and Domingos, ``Markov Logic Networks,'' \emph{Machine Learning}, 2006) combine first-order logic with Markov random fields. Like the QBBN, MLNs weight logical formulas and perform probabilistic inference. Unlike the QBBN, MLNs handle full first-order logic formulas---not just Horn clauses---using undirected graphical models with inference via MCMC sampling. The QBBN restricts expressiveness to the forward fragment of natural deduction (universally quantified Horn clauses with negation) but gains efficient inference via belief propagation rather than MCMC. The boolean decomposition into AND, OR, and NEG factors enables tractable message passing precisely because the forward fragment avoids the case analysis and witness search that make full first-order inference intractable. Additionally, MLNs do not address the parsing problem---they assume logical forms are given.

\subsection{ProbLog}
ProbLog (De Raedt, Kimmig, and Toivonen, ``ProbLog: A Probabilistic Prolog and Its Application in Link Discovery,'' IJCAI, 2007) extends Prolog with probabilistic facts. ProbLog uses exact inference via weighted model counting, which is theoretically complete but computationally expensive for large knowledge bases. The QBBN trades exactness for efficiency through approximate belief propagation, and uses a role-based predicate language rather than Prolog's positional arguments.

\subsection{Neural Theorem Provers}
Rockt\"aschel and Riedel (``End-to-End Differentiable Proving,'' NeurIPS, 2017) introduced neural theorem provers that learn to perform logical inference using differentiable unification. These systems can learn rules from data but struggle with the precision required for formal logical reasoning. Our approach separates learning (in the LLM preprocessing layer) from reasoning (in the QBBN inference engine), achieving exactness in the reasoning component.

\subsection{Neuro-Symbolic Approaches}
The broader neuro-symbolic literature seeks to combine neural networks with symbolic reasoning. Two representative works illustrate the dominant paradigm and its contrast with our approach.

Garcez, Lamb, and Gabbay (\emph{Neural-Symbolic Cognitive Reasoning}, Springer, 2009) proposed encoding logical rules directly into neural network connection weights, so that a neural network's forward pass performs logical inference. Their system covers propositional, modal, temporal, and epistemic logics, and can learn from data while maintaining logical structure. However, once logic is embedded in network weights, the reasoning becomes opaque---the system cannot explain its inferences in terms of explicit causal chains, unlike the QBBN's factor graph, where every conclusion traces back through named rules and typed propositions.

Manhaeve et al.\ (``DeepProbLog: Neural Probabilistic Logic Programming,'' NeurIPS, 2018) extended ProbLog with \emph{neural predicates}---predicates whose truth probabilities are computed by neural networks. For example, a convolutional network classifies MNIST digit images, and a logic program defines addition over the classified digits. The entire system is trained end-to-end via gradient semirings, backpropagating through both the logic program and the neural network.

Both of these approaches, and the differentiable reasoning paradigm more broadly, seek to make logic \emph{differentiable}: they embed discrete logical operations into continuous, gradient-friendly frameworks so that the entire system can be trained end-to-end by backpropagation. This is an elegant strategy for learning, but it comes at a cost. Differentiable approximations of discrete logical operations introduce noise, and the resulting systems cannot guarantee the exactness of their logical conclusions.

Our architecture makes a fundamentally different choice. We impose a clean separation between the neural component (the LLM) and the symbolic component (the grammar and the QBBN). There is no differentiable path from the QBBN's inference output back to the LLM's weights. The LLM produces discrete symbolic output---disambiguated text---which the grammar compiles to discrete logical forms, which the QBBN reasons over using exact boolean factors and belief propagation. This separation means we cannot train the system end-to-end, which is a real limitation for learning from data. But it means that the reasoning component is exact: when the QBBN concludes $P(p) > 0.5$, that conclusion follows from explicit, inspectable logical derivations, not from a differentiable approximation of logic. We view this as a necessary property for a system intended to reason without hallucinating.

Neither Garcez et al.\ nor DeepProbLog addresses the problem of parsing natural language to logical form. Their logical forms are given as input, either as hand-specified rules or as structured queries. Our typed lexicon serves as the interface between the neural and symbolic worlds, providing a principled boundary that neither the differentiable approaches nor the classical symbolic approaches have articulated.

\subsection{Natural Deduction and Proof Theory}
Our system is grounded in the natural deduction tradition (Gentzen, 1934; Prawitz, 1965). The boolean decomposition into AND, OR, and NEG factors implements the simple elimination rules of natural deduction, restricted to the forward fragment. This fragment corresponds to universally quantified Horn clauses with negation---the same fragment studied in Datalog and logic programming, but realized here as a probabilistic factor graph rather than a resolution-based prover. The key difference from classical logic programming is that our system uses belief propagation rather than backtracking search, and produces probabilities rather than yes/no answers.

Prawitz (1965) provides the theoretical foundation for the system's expressiveness claims. His treatment of second-order logic (Chapter V) grounds the extension to predicate quantification and lambda abstraction. His treatment of modal logic (Chapter VI) grounds the extension to propositional attitudes and intensional verbs. The fragment hierarchy---forward (linear time), query (existential search), planning (NP-hard), full (undecidable)---provides a principled account of why the current system is tractable and where future extensions sit in the complexity landscape.

\subsection{Semantic Parsing}
The compositional semantic parsing tradition, from Montague (``The Proper Treatment of Quantification in Ordinary English,'' 1973) through Zettlemoyer and Collins (``Learning to Map Sentences to Logical Form,'' UAI, 2005) to modern neural semantic parsers, addresses the same text-to-logic problem. Our typed slot grammar is a deliberately simple parser that trades coverage for exactness---it handles fewer constructions than a CCG parser (Steedman, \emph{The Syntactic Process}, MIT Press, 2000) but guarantees zero ambiguity on the constructions it does handle. The LLM preprocessing layer is designed to extend coverage without sacrificing this guarantee.

\subsection{LLMs as Linguists}
Recent work has shown that LLMs possess substantial metalinguistic knowledge. Blevins, Gonen, and Zettlemoyer (``Prompting Language Models for Linguistic Structure,'' ACL, 2023) showed that LLMs can output structured linguistic analyses with appropriate prompting. Gulordava et al.\ (``Colorless Green Recurrent Networks Dream Hierarchically,'' NAACL, 2018) demonstrated that language models capture long-distance syntactic dependencies. Ettinger et al.\ (``You Are an Expert Linguistic Annotator: Limits of LLMs as Analyzers of Abstract Meaning Representation,'' arXiv:2310.17793, 2023) found that while LLMs can reproduce AMR format, they fail to generate semantically accurate parses---consistent with our finding that LLMs understand structure but cannot produce exact formal representations.

%% file: sections/11_conclusion.tex
\section{Conclusion}
\label{sec:conclusion}

The overall project contributes a logical Bayesian network---a system that unifies logical and probabilistic inference by decomposing weighted Horn clauses into AND, OR, and NEG factors on a bipartite graph and running Pearl-style belief propagation. Prior probabilistic logic systems used weighted model counting (ProbLog) or MCMC sampling (Markov Logic Networks). The boolean decomposition with belief propagation is, to our knowledge, new.

This paper makes three contributions corresponding to the three classical areas of natural language processing. For \textbf{inference}, we introduce the NEG factor and bidirectional proposition graph construction, extending the QBBN from forward-only reasoning to the full forward fragment of natural deduction (Prawitz, 1965)---both modus ponens and modus tollens within a single belief propagation loop. For \textbf{semantics}, we present a typed logical language with three tiers of expressiveness following Prawitz---first-order quantification over entities, propositions as arguments, and a designed extension to predicate quantification---and argue that these tiers are sufficient for natural language semantics. For \textbf{syntax}, we present a grammar-first, agent-assisted architecture in which a deterministic grammar provides exactness and LLM-assisted disambiguation provides coverage.

\subsection*{The Bitter Lesson, Revisited}

Sutton (2019) argued that hand-engineered knowledge representations inevitably lose to general methods that scale with computation. The formal semantics tradition---from Montague through Steedman to the present work---was historically on the losing side of this argument, because the bottleneck was human annotation: every grammar rule, lexicon entry, and coverage test required a trained linguist. The representation was sound but the construction process did not scale.

We argue that large language models eliminate this bottleneck. The LLM serves as the annotator---generating grammar rules, lexicon entries, logical forms, and coverage tests---while the formal inference engine serves as the verifier. This division aligns with the bitter lesson's prescription: both the annotation (LLM, scaling with model size and training data) and the inference (belief propagation, scaling with graph size and iteration count) scale with computation. What no longer scales with human effort is the construction of the formal infrastructure, because the LLM replaces the human in that role.

The system described in this paper was built in a semi-automatic bridge stage using the ``vibe coding'' pattern (Karpathy, 2025)---LLM-assisted construction of code, experiments, and linguistic annotations, with the human providing direction and quality judgment. Earlier iterations were developed with ChatGPT (OpenAI); the current implementation and paper were developed primarily with Claude (Anthropic). Both LLMs contributed to code development, test generation, and paper drafting. This bridge is explicitly temporary: once sufficient grammar coverage exists, the LLM produces logical forms directly and the QBBN verifies them, with no human in the loop.

\subsection*{Next Steps}

The immediate next steps are extending grammar coverage to the full set of 22 reasoning patterns tested by the inference engine (the grammar currently covers 12), scaling to larger knowledge bases, and evaluating the end-to-end pipeline on open-domain text. Longer-term goals include learning rule weights from data via the EM algorithm, integrating the system as a verification layer for LLM-generated reasoning, and demonstrating the five capabilities outlined in Section~\ref{sec:motivation}---hallucination-free inference, verifiable reasoning, goal-directed planning, continuous learning, and explicit world models---at scale.

The logical and inferential machinery is in place. What remains is lexical: determining which words map to which predicates and types, and demonstrating this at scale. The bitter lesson tells us this is precisely the kind of work that should be done by computation, not by hand. We agree, and the LLM is the computation.

All code, test data, extended documentation, and inference engine implementation are open source at \url{https://github.com/gregorycoppola/world}.

%% file: sections/11b_acknowledgments.tex
\section*{Acknowledgments}

This work was developed with substantial assistance from large language models, a fact we acknowledge both for transparency and as evidence for the paper's central argument that LLMs can serve as annotators and collaborators in formal linguistic research.

The 2024 paper and initial codebase (February 2024 through approximately December 2025) were developed primarily with ChatGPT (OpenAI), which assisted with code generation, debugging, and early explorations of the logical language design. The current implementation, coverage tests, grammar rules, and this paper (December 2025 through February 2026) were developed primarily with Claude (Anthropic), which contributed to code development, test generation, grammar authoring, inference engine debugging, and paper drafting and review.

Both LLMs reviewed the logical representations, checked inference results against expected outputs, and critiqued the paper's argumentation---providing a form of automated review that complements traditional peer review. The fact that two independent LLM systems, trained on different data by different organizations, could each engage productively with the formal representations described here is itself evidence that the representations are well-defined and machine-legible.

We encourage readers to verify our claims by loading the open-source code and documentation into any capable LLM. The logical language, grammar rules, coverage tests, and inference engine are designed to be legible to both humans and language models. The repository is available at \url{https://github.com/gregorycoppola/world}.

%% file: sections/12_bibliography.tex
\section*{References}

\begin{description}

\item[Ambridge and Blything, 2024] Ambridge, B.\ and Blything, L.\ ``Large Language Models Are Better Than Theoretical Linguists at Theoretical Linguistics.'' \emph{Theoretical Linguistics}, 50:33--48, 2024.

\item[Bar-Hillel, 1953] Bar-Hillel, Y.\ ``A Quasi-Arithmetical Notation for Syntactic Description.'' \emph{Language}, 29(1):47--58, 1953.

\item[Beg\v{u}s et al., 2025] Beg\v{u}s, G., Dabkowski, M., and Rhodes, R.\ ``Large Linguistic Models: Investigating LLMs' Metalinguistic Abilities.'' arXiv:2305.00948, 2025.

\item[Blevins et al., 2023] Blevins, T., Gonen, H., and Zettlemoyer, L.\ ``Prompting Language Models for Linguistic Structure.'' In \emph{Proceedings of ACL}, pages 6649--6663, 2023.

\item[Chomsky, 1957] Chomsky, N.\ \emph{Syntactic Structures}. Mouton, 1957.

\item[Chomsky, 1965] Chomsky, N.\ \emph{Aspects of the Theory of Syntax}. MIT Press, 1965.

\item[Cooper, 1990] Cooper, G.F.\ ``The Computational Complexity of Probabilistic Inference Using Bayesian Belief Networks.'' \emph{Artificial Intelligence}, 42(2--3):393--405, 1990.

\item[Coppola, 2024] Coppola, G.\ ``The Quantified Boolean Bayesian Network: Theory and Experiments with a Logical Graphical Model.'' arXiv:2402.06557, 2024.

\item[De Raedt et al., 2007] De Raedt, L., Kimmig, A., and Toivonen, H.\ ``ProbLog: A Probabilistic Prolog and Its Application in Link Discovery.'' In \emph{Proceedings of IJCAI}, 2007.

\item[Dempster et al., 1977] Dempster, A.P., Laird, N.M., and Rubin, D.B.\ ``Maximum Likelihood from Incomplete Data via the EM Algorithm.'' \emph{JRSS-B}, 39(1):1--38, 1977.

\item[Eisner, 1996] Eisner, J.\ ``Three New Probabilistic Models for Dependency Parsing.'' In \emph{Proceedings of COLING}, pages 340--345, 1996.

\item[Ettinger et al., 2023] Ettinger, A., Hwang, J.D., Pyatkin, V., Bhagavatula, C., and Choi, Y.\ ``You Are an Expert Linguistic Annotator: Limits of LLMs as Analyzers of Abstract Meaning Representation.'' arXiv:2310.17793, 2023.

\item[Garcez et al., 2009] Garcez, A.S., Lamb, L.C., and Gabbay, D.M.\ \emph{Neural-Symbolic Cognitive Reasoning}. Springer, 2009.

\item[Gentzen, 1934] Gentzen, G.\ ``Untersuchungen \"uber das logische Schlie\ss en.'' \emph{Mathematische Zeitschrift}, 39:176--210, 405--431, 1934.

\item[Gormley et al., 2015] Gormley, M.R., Dredze, M., and Eisner, J.\ ``Approximation-Aware Dependency Parsing by Belief Propagation.'' \emph{TACL}, 3:489--501, 2015.

\item[Gulordava et al., 2018] Gulordava, K., Bojanowski, P., Grave, E., Linzen, T., and Baroni, M.\ ``Colorless Green Recurrent Networks Dream Hierarchically.'' In \emph{Proceedings of NAACL}, pages 1195--1205, 2018.

\item[Karpathy, 2025] Karpathy, A.\ ``Vibe Coding.'' Post on X (formerly Twitter), February 2, 2025. \url{https://x.com/karpathy/status/1886192184808149383}.

\item[Koeva, 2024] Koeva, S.\ ``Large Language Models in Linguistic Research: The Pilot and the Copilot.'' In \emph{Proceedings of CLIB}, pages 319--328, 2024.

\item[Koller and Friedman, 2009] Koller, D.\ and Friedman, N.\ \emph{Probabilistic Graphical Models: Principles and Techniques}. MIT Press, 2009.

\item[Manhaeve et al., 2018] Manhaeve, R., Dumancic, S., Kimmig, A., Demeester, T., and De Raedt, L.\ ``DeepProbLog: Neural Probabilistic Logic Programming.'' In \emph{Proceedings of NeurIPS}, 2018.

\item[McDonald et al., 2005] McDonald, R., Pereira, F., Ribarov, K., and Haji\v{c}, J.\ ``Non-Projective Dependency Parsing Using Spanning Tree Algorithms.'' In \emph{Proceedings of HLT-EMNLP}, pages 523--530, 2005.

\item[Montague, 1970] Montague, R.\ ``Universal Grammar.'' \emph{Theoria}, 36:373--398, 1970.

\item[Montague, 1973] Montague, R.\ ``The Proper Treatment of Quantification in Ordinary English.'' In \emph{Approaches to Natural Language}, pages 221--242, 1973.

\item[Murphy et al., 1999] Murphy, K., Weiss, Y., and Jordan, M.I.\ ``Loopy Belief Propagation for Approximate Inference: An Empirical Study.'' In \emph{Proceedings of UAI}, pages 467--476, 1999.

\item[Neapolitan, 2003] Neapolitan, R.E.\ \emph{Learning Bayesian Networks}. Prentice Hall, 2003.

\item[Pearl, 1988] Pearl, J.\ \emph{Probabilistic Reasoning in Intelligent Systems: Networks of Plausible Inference}. Morgan Kaufmann, 1988.

\item[Prawitz, 1965] Prawitz, D.\ \emph{Natural Deduction: A Proof-Theoretical Study}. Almqvist \& Wiksell, 1965.

\item[Richardson and Domingos, 2006] Richardson, M.\ and Domingos, P.\ ``Markov Logic Networks.'' \emph{Machine Learning}, 62:107--136, 2006.

\item[Rockt\"aschel and Riedel, 2017] Rockt\"aschel, T.\ and Riedel, S.\ ``End-to-End Differentiable Proving.'' In \emph{Proceedings of NeurIPS}, 2017.

\item[Roth, 1996] Roth, D.\ ``On the Hardness of Approximate Reasoning.'' \emph{Artificial Intelligence}, 82:273--302, 1996.

\item[Smith and Eisner, 2008] Smith, D.\ and Eisner, J.\ ``Dependency Parsing by Belief Propagation.'' In \emph{Proceedings of EMNLP}, pages 145--156, 2008.

\item[Steedman, 1996] Steedman, M.\ \emph{Surface Structure and Interpretation}. MIT Press, 1996.

\item[Steedman, 2000] Steedman, M.\ \emph{The Syntactic Process}. MIT Press, 2000.

\item[Sutton, 2019] Sutton, R.\ ``The Bitter Lesson.'' Blog post, March 13, 2019. \url{http://www.incompleteideas.net/IncIdeas/BitterLesson.html}.

\item[Zettlemoyer and Collins, 2005] Zettlemoyer, L.S.\ and Collins, M.\ ``Learning to Map Sentences to Logical Form: Structured Classification with Probabilistic Categorial Grammars.'' In \emph{Proceedings of UAI}, pages 658--666, 2005.

\end{description}